\documentclass[preprint,12pt]{elsarticle}

\usepackage[margin=1.25in,footskip=0.5in]{geometry}
\usepackage{color,soul}

\usepackage{booktabs}
\usepackage{amssymb}
\usepackage{comment}
\usepackage{xcolor}
\usepackage{ulem}
\DeclareSymbolFont{largesymbol}{OMX}{yhex}{m}{n}

\usepackage[export]{adjustbox}
\usepackage{amsmath}
\usepackage{float}

\journal{Neurocomputing}

\usepackage[bookmarks,bookmarksnumbered]{hyperref}
\hypersetup{colorlinks = true,linkcolor = green,anchorcolor =red,citecolor = green,filecolor = red,urlcolor = red, pdfauthor=author}

\graphicspath{{figure/}}

\begin{document}

\begin{frontmatter}

\title{ \large A Survey on Epistemic (Model) Uncertainty in Supervised Learning: Recent Advances and Applications}

\author[First]{Xinlei Zhou}
\ead{Zhouxnli@gmail.com}

\author[First,Second]{Han Liu}
\ead{han.liu@szu.edu.cn}

\author[Second,Third]{Farhad Pourpanah}
\ead{farhad@szu.edu.cn,farhad.086@gmail.com}

\author[Fourth]{Tieyong Zeng}
\ead{zeng@math.cuhk.edu.hk}
 
\author[First,Second]{Xizhao Wang\corref{cor}}
\ead{xizhaowang@ieee.org}

\cortext[cor]{Corresponding author}

\address[First]{\scriptsize College of Computer Science and Software Engineering, Shenzhen University, Shenzhen 518060, China}
\address[Second]{The Guangdong Key Laboratory of Intelligent Information Processing, Shenzhen University, 518060, China}
\address[Third]{College of Mathematics and Statistics, Shenzhen University, Shenzhen 518060, China}
\address[Fourth]{ Department of Mathematics, The Chinese University of Hong Kong, Hong Kong}

\begin{abstract}
Quantifying the uncertainty of supervised learning models plays an important role in making more reliable predictions. \textit{Epistemic} uncertainty, which usually is due to insufficient knowledge about the model, can be reduced by collecting more data or refining the learning models. Over the last few years, scholars have proposed many epistemic uncertainty handling techniques which can be roughly grouped into two categories, i.e., Bayesian and ensemble. This paper provides a comprehensive review of epistemic uncertainty learning techniques in supervised learning over the last five years. As such, we, first, decompose the epistemic uncertainty into bias and variance terms. Then, a hierarchical categorization of epistemic uncertainty learning techniques along with their representative models is introduced. In addition, several applications such as computer vision (CV) and natural language processing (NLP) are presented, followed by a discussion on research gaps and possible future research directions.         

\end{abstract}


\end{frontmatter}

\section{Introduction}
\label{sec:intro}
Supervised learning, as a broad branch of machine learning, refers to the task of learning a mapping function for associating high-dimensional input samples into their corresponding target vectors using labeled data~\cite{wang2020recent,pourpanah2020review,rezvani2019intuitionistic,lou2021dual}. They have been successfully used for a variety of real-world applications, e.g., medical and fault diagnosis~\cite{garg2018perfect,nair2020exploring,wang2021fuzzy,pourpanah2018anomaly}, object detection~\cite{he2019bounding, kraus2019uncertainty}, text processing~\cite{xiao2019quantifying,Wang:NLP,He:NLP}, and speech recognition~\cite{daubener2020detecting}{\color{
black}, image segmentation~\cite{badrinarayanan2017segnet, zeng2021deep, zeng2019improved}, image enhancement~\cite{fang2020soft,liu2021melt}}. 
Indeed, supervised learning is a process of predicting unknown data based on partial samples that cannot accurately represent the whole data set distribution. In such an experience-driven process, the model is not provably correct but only hypothetical; therefore uncertain and the same holds for the predictions produced by the model~\cite{hullermeier2021aleatoric}. In addition, the challenge of big data, such as skyrocketed feature dimensions and categories, missing data, unbalanced data distribution and huge solution space, aggravate the uncertainty of the learning process, which seriously affects the performance of the supervised learning algorithms~\cite{wang2016learning}. Moreover, supervised learning approaches are unable to identify in-domain from out-domain samples~\cite{ovadia2019can}, provide reliable uncertainty approximation~\cite{ayhan2018test}, and lack expressiveness during inference~\cite{guha2019bayesian}; therefore, their deployment in high-risk and safety-critical applications remains limited. To alleviate these issues, it is vital to present uncertainty estimate in a way that ignores the uncertain predictions or passes them to experts~\cite{shen2021study}.


{\color{black}In supervised learning, traditional uncertainty assessment is usually based on a single probability distribution. Nowadays, the widely accepted way is quantifying uncertainty separately by distinguishing two different sources, i.e., \textit{aleatoric} uncertainty and \textit{epistemic} uncertainty~\cite{hullermeier2021aleatoric}. Aleatoric (data) uncertainty is a kind of uncertainty that reflects the inherent property of data, like noise. It is usually caused by the irreducible error in the data measurements and observations process, which cannot be reduced even by collecting more data. Kendal and Gal~\cite{kendall2017uncertainties} further divided aleatoric uncertainty into \textit{homoscedastic} uncertainty and \textit{heteroscedastic} uncertainty. The former is a value that stays constant for various input samples in the same task and the latter is associated with the differences among input data, for example, some inputs contain more noise than others.}
In contrast, epistemic (model) uncertainty is referred to a state that model cognition is restricted, which is due to the upper limit of the model fitting ability, the optimizing strategy, the parameters, the lack of knowledge. It can be reduced by gathering more data or refining models.

On the other hand, the generalization error is a standard metric to quantify the effectiveness of decisions made by supervised learning models. Meanwhile, several studies~\cite{wang2014study,wang2017discovering,zhou2020analysis} have proved that the generalization error manifests the predictive uncertainty. It simultaneously commits to the theoretical exploration of the quantification and formal expression of their relationship.
The generalization error can be decomposed into three terms, i.e., noise, bias, and variance~\cite{GAO2021100309,friedman2001elements}.

\begin{figure}[tb!]
\centering
\includegraphics[width=0.9\linewidth]{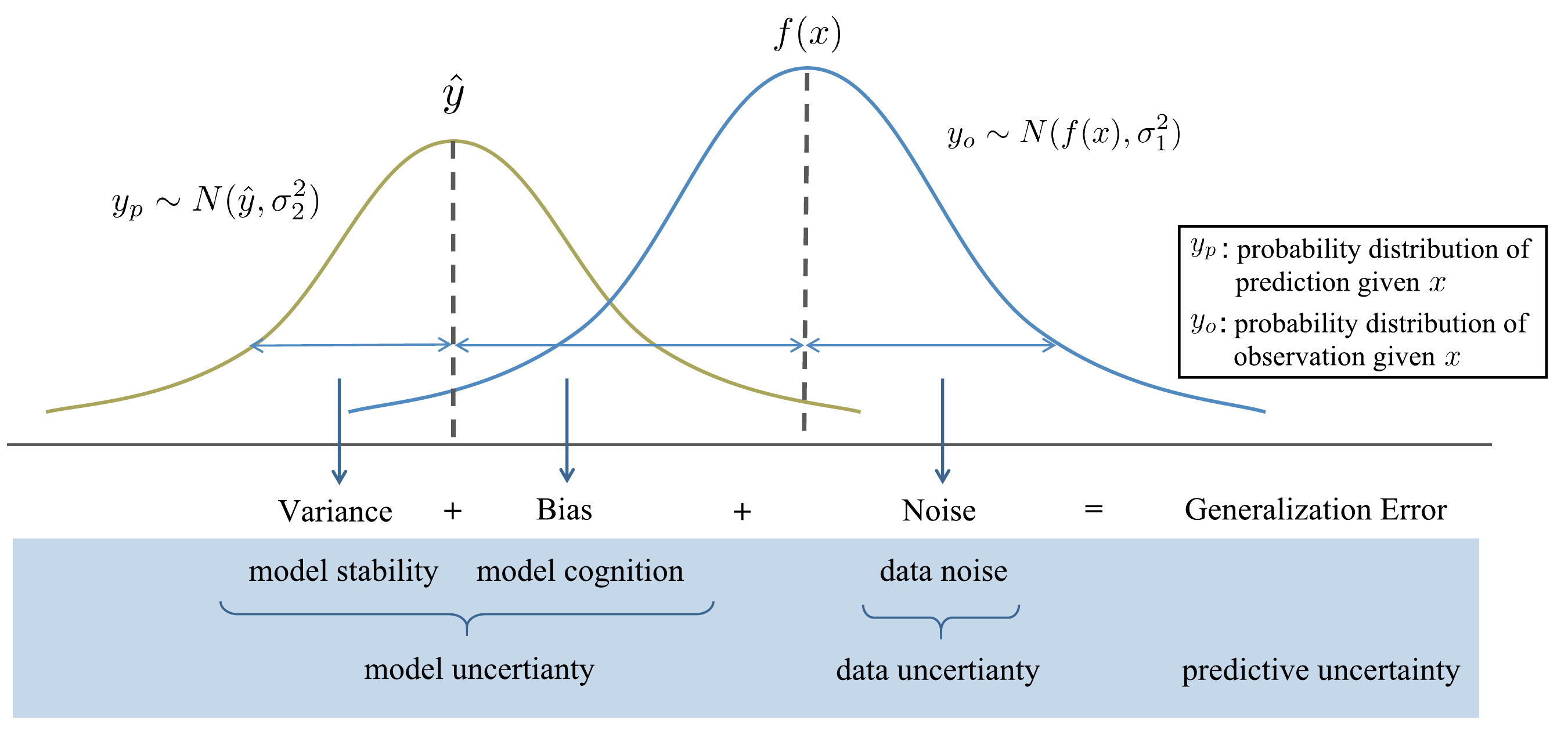}
\caption{ Decomposition of uncertainty in supervised learning.}
\label{Fig:dec}
\end{figure}

Suppose, $y_o = f(\textbf{x}) + \epsilon$ represents the observed value for a given $\textbf{x}\in \Re^{d}$, which is corrupted by noise $\epsilon \sim N(0,\sigma_1^2)$. 
Thus, $y_o \sim N(f(\textbf{x}),\sigma_1^2)$ where $f(\textbf{x})$ represents the original target. 
Besides, $y_p \sim N(\hat{y},\sigma_2^2)$ denotes the distribution of prediction $\hat{f}(\textbf{x})$ centered on mean value $\hat{y}=\mathbb{E}\hat{f}(\textbf{x})$. 
The noise term ($\sigma_1^2$) arises from data and it is irreducible, which can be represented as aleatoric (data) uncertainty. In contrast, the (squared) bias term ($[\mathbb{E}\hat{f}(\textbf{x})-f(\textbf{x})]^2$) reveals the gap between the estimated value and the true value. It reflects the degree of cognitive limitation caused by the setting of model properties such as parameters, strategies, or learning algorithms. While, the variance term ($\mathbb{E}[\hat{f}(\textbf{x})-\mathbb{E}\hat{f}(\textbf{x})]^2$) is related to the sensitivity of model pertaining to the training samples. Thus, we argue that the bias term together with variance represents the epistemic (model) uncertainty.
Fig.~\ref{Fig:dec} shows these three terms and explains the predictive uncertainty from the perspective of the generalization error decomposition. The generalization error of a model can be reduced through the correlation analysis of bias and variance, i.e., epistemic uncertainty. 
Therefore, analyzing the epistemic uncertainty, using the established relationship between uncertainty and error items (as shown in Fig.~\ref{Fig:dec}), can help to select an appropriate uncertainty quantification method and improve the model performance.

\begin{table}[tb!]
\small
\renewcommand{\arraystretch}{1}
\caption{Summary of related uncertainty quantification surveys.}
\label{table:listreview}
\begin{adjustbox}{width=1\linewidth}
    \begin{tabular}{p{4cm}p{3.5cm}p{5cm}}
    \toprule
    Study & Venue & Content\\
    \midrule
    Wang and He (2016)~\cite{wang2016learning} & IEEE Systems, Man, \& Cybernetics Magazine & Summarizing challenges of big data and uncertainty-based learning methods.\\
    \midrule
    Kabir et al. (2018)~\cite{kabir2018neural}& IEEE access & Discussing from the concept of Prediction Intervals.\\
    \midrule
    Hariri et al. (2019)~\cite{hariri2019uncertainty}& Journal of Big Data & Categorizing techniques that handle uncertainty in big data according to various data characteristics.\\
    \midrule
    Wang and Yeung (2020)~\cite{wang2020survey}&  ACM Computing Surveys & Outlining Bayesian-based quantification methods from the perspective of PGM.\\
    \midrule
    Jospin et al. (2020)~\cite{jospin2020hands}&  ACM Computing Surveys & A tutorial specific on Bayesian deep learning.\\
    \midrule
    Hullermeier and  Waegeman (2021)~\cite{hullermeier2021aleatoric}& Machine Learning & Emphasizing the important to distinguish different uncertainty.\\
    \midrule
    Abdar et al. (2021)~\cite{abdar2021review}&  Information Fusion & Reviewing uncertainty quantification techniques along with their applications.\\
    \midrule
    Gawlikowski et al. (2021)~\cite{gawlikowski2021survey}& arXiv & Introducing sources of uncertainty, categorizing uncertainty techniques and reviewing re-calibration techniques .\\
    \bottomrule
    \end{tabular}
\end{adjustbox}
\end{table}





\textbf{Existing survey papers}:
There exist several overviews of uncertainty learning techniques in machine learning from different perspectives and emphases (see Table~\ref{table:listreview}).  In 2016, Wang and He~\cite{wang2016learning} discussed the challenging issues in analyzing big data and emphasized the importance of modeling uncertainty in improving the performance of the learning models. Subsequently, Hariri et al.~\cite{hariri2019uncertainty} surveyed uncertainty learning techniques in big data. They briefly introduced the classical uncertainty measuring techniques in machine learning and categorized them into probability theory, Shannon's entropy, fuzzy set theory, and rough set theory. Recently, Hullermeier and Waegeman~\cite{hullermeier2021aleatoric} emphasized the significance of identifying aleatoric and epistemic uncertainty separately in machine learning. With the popularity of deep learning techniques, most of the recent review papers focused on techniques that are effective for neural network frameworks. For example, Kabir et al.~\cite{kabir2018neural} provided a review of uncertainty quantification techniques in neural networks from the concept of prediction intervals. Wang et al.~\cite{wang2020survey} described Bayesian deep learning as a uniform framework that combines deep learning techniques with a paradigm of excellent uncertainty handling capabilities, i.e., probabilistic graphical methods. Jospin et al.~\cite{jospin2020hands} organized a handbook from basic statistic concepts to the principle, the learning strategy, and specific algorithms for researchers interested in Bayesian neural networks. They categorized Bayesian methods into Variational inference (VI), Markov Chain Monte Carlo (MCMC), and Bayes by backprop. They also discussed approximation techniques in terms of stochastic gradient descent (SGD) dynamics and Monte Carlo dropout (MCD). 

Recently, Abdar et al.~\cite{abdar2021review} gave an extensive review of uncertainty quantification methods in deep learning along with their applications. Specifically, they categorized the uncertainty quantification techniques into Bayesian approximation and ensemble learning and discussed the representative models of each category. In addition, they provided open challenges and future research directions associated with uncertainty quantification. While Gawlikowski et al.~\cite{gawlikowski2021survey} first identified five sources of uncertainty in deep learning models. These sources are variability in practical scenarios, error, and noise in measurement tools, errors caused by unknown data samples, errors in model structure and training procedure. Then, they categorized uncertainty learning techniques into single deterministic methods, Bayesian methods, ensemble Methods, test-time augmentation methods. Besides, they reviewed re-calibration techniques in DL. Each of these surveys has provided a comprehensive review of uncertainty learning from different points of view. But none of them include a detailed review of epistemic uncertainty techniques in terms of bias-variance decomposition.

\textbf{Contributions of the paper}:
Different from existing surveys, we emphasize the importance of uncertainty analysis in supervised learning models from the perspective of generalization error decomposition. Specifically, the focus is on tracing the epistemic uncertainty according to the decomposed items, i.e., bias and variance. In this paper, we outline the two most representative techniques in supervised learning, e.g., Bayesian and ensemble methods, over the last five years and discuss the properties of each category according to the bias-variance decomposition. In this context, we have collected a total of 138 publications, in which:
\begin{itemize}
    \item {\color{black}Approximately 70\% of the selected articles are published in the last five years, i.e., after 2016; while most of the remaining 30\% are high-cited classic articles.}
    \item More than 80\% of the selected articles are indexed in Q1 journals, top conferences\footnote{Top conferences refer to the well-accepted high level conferences, such as NIPS, ICML, AAAI, IJCAI, ICCV, ECCV, CVPR, etc.}, and high-cited books or thesis 
\end{itemize}

Standing upon these high-quality articles, this work aims to guide researchers interested in tracing the limitation of models from the perspective of uncertainty decomposition, quantification, analysis, and applications. 
To sum up, the main contributions of this survey include: 

\begin{itemize}
    \item review of the epistemic (model) uncertainty learning techniques in supervised models over the last five years;
    \item {\color{black}discussion on epistemic uncertainty learning from the perspective 
    of generalization error, i.e., bias and variance decomposition; } 
    \item hierarchical categorization of the epistemic uncertainty learning methods along with their representative models and real-world applications;
    \item elucidation on the main research gaps and suggesting future research directions.
\end{itemize}


\begin{figure}[tb!]
\centering
\includegraphics[width=1\linewidth]{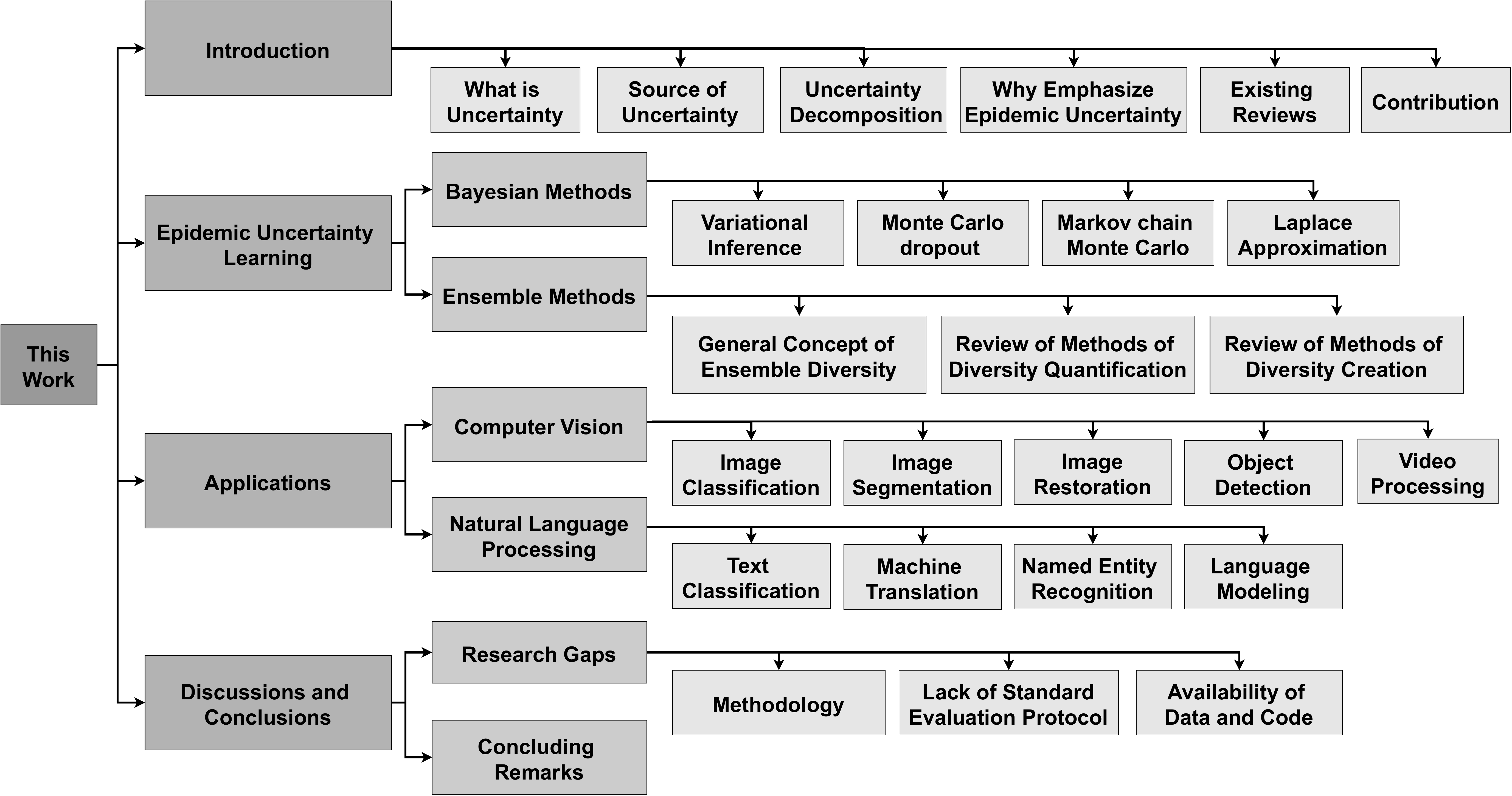}
\caption{Organization of this survey.}
\label{Fig:org}
\end{figure}
 
\textbf{Organization of the paper}:
As shown in Fig.~\ref{Fig:org}, this review consists of four sections. Section~\ref{sec:review},  first, provides a review of epistemic uncertainty learning methods. Specifically, these methods are categorized into Bayesian approximation and ensemble learning. Then, several widely used Bayesian approximation techniques, such as variational inference (VI), Monte Carlo dropout (MCD), Markov Chain Monte Carlo (MCMC), and Laplace approximation (LA), are discussed in detail. Finally, ensemble learning is introduced in terms of its concept and relationship to epistemic uncertainty as well as those related ensemble methods.
Section~\ref{sec:app} discusses the importance of quantifying uncertainty in supervised learning approaches for several real-world applications such as computer vision and natural language processing. Section~\ref{sec:discuss} presents the research gaps and trends for future research as well as concluding remarks.

\section{Review on Epistemic Uncertainty Learning}
\label{sec:review}

\begin{figure}[tb!]
\centering
\includegraphics[width=0.9\linewidth]{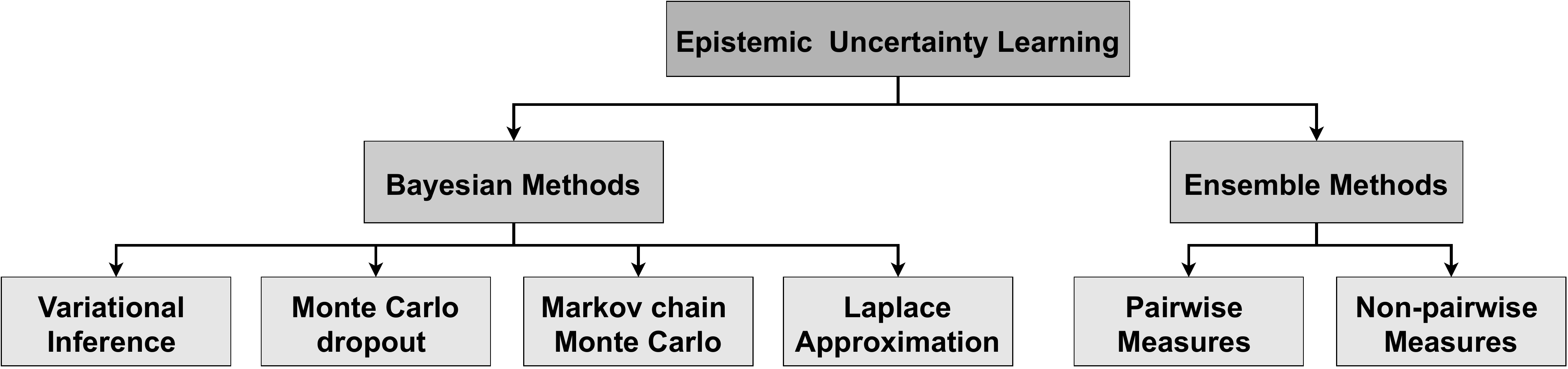}
\caption{ The taxonomy of epistemic (model) uncertainty learning.}
\label{Fig:taxo}
\end{figure}

This section gives a review of the epistemic uncertainty learning techniques in supervised learning. Specifically, we focus on the epistemic uncertainty quantification methods in terms of decomposed items of the generalization error, i.e., bias and variance. According to Fig.~\ref{Fig:taxo}, the epistemic uncertainty learning methods are grouped into the Bayesian and ensemble methods, as follows:

{\color{black}
\begin{itemize}
\item \textbf{Bayesian methods} formulate epistemic uncertainty as a probability distribution over the model parameters. These methods mainly quantify the variance (as shown in Fig.~\ref{Fig:dec}) in order to reduce the generalization error of the learning model. In this context, most studies explore the neural network-based model and Bayesian methods to estimate the epistemic uncertainty caused by the model parameters. 
Section~\ref{sec:sec:bayes} provides a comprehensive review of these techniques. 
\item \textbf{Ensemble methods} train multiple models to produce multiple predictions and then combine their predictions to reach the final output. These methods mainly quantify the variance of the outputs of base models as the epistemic uncertainty to control the complementarity among these base models for improving the ensemble performance. In this context, the reduction of the generalization error can be achieved by reducing the bias or the variance of the ensemble output, depending on the essence of specific ensemble methods. These methods are reviewed in Section~\ref{sec:sec:ensemble}.
\end{itemize}
}
In the following subsections, we provide a detailed review and discuss the main properties of each category.

\subsection{Bayesian methods}
\label{sec:sec:bayes}
Assume a training data set $D=\{(\textbf{x}_i,y_i)\}_{i=1}^{N}$, where $\textbf{x}_i \in \Re^{d}$ and $y_i \in \{1,...,C\}$ indicate the $i$-th input and its corresponding class, respectively, and $C$ denotes the number of classes.
The aim is to learn a function $y=f^{\theta}(\textbf{x})$ with parameters $\theta$ to obtain a desired output.
Bayesian modeling aims to capture the epistemic uncertainty by putting distributions over the network weights instead of deterministic network weights, which is known as \textit{marginalisation}.
For a given test sample \textbf{x}$^*$, the distribution over a prediction $y^*$ can be written as~\cite{gal2016uncertainty}:
\begin{align}
\label{eq:predictivedistribution}
p(y^*|\textbf{x}^*,D) = \int p(y^*|\textbf{x}^*,\theta)
p(\theta|D) d\theta,
\end{align}
where $p(\theta|D)$, which is known as posterior distribution on the model parameters, represents the uncertainty on the model parameters given a training data set $D$.

Assume $\widehat{\theta}_t$ indicates the parameters of $t$-th sample from distribution $p(\theta|D)$, the epistemic uncertainty of the model can be quantified via variance term (as shown in Fig.~\ref{Fig:dec}):
\begin{align}
\label{eq:bayesian_var}
Var(y*) \approx \frac{1}{T} \sum_{t=1}^T f^{\widehat{\theta}_t}(\textbf{x}^*)^\mathsf{T}f^{\widehat{\theta}_t}(\textbf{x}^*)-\mathbb{E}(y*)^\mathsf{T}\mathbb{E}(y*),
\end{align}
where $T$ is the total number of sampling, which will be explained in the following subsections.

Bayesian methods are easy to implement but difficult to perform inference, because they require to estimate the posterior distribution, i.e., $p(\theta |D)$. Therefore, the marginal probability cannot be computed analytically~\cite{kendall2017uncertainties}. In order to obtain the posterior distribution, the Bayes theorem~\cite{anzai2012pattern} is applied to a given data set $D$ over $\theta$, as follows: 
\begin{align}
\label{eq:post}
p(\theta |D)=\frac{p(D,\theta)p(\theta)}{p(D)},
\end{align}
{\color{black}where $p(D,\theta)$ represents the likelihood that the data samples in $D$ are 
realization of the distribution predicted by a model with parameter $\theta$, and $p(D)$ is the prior distribution on the model parameters.} 
Scholars have proposed many approximation techniques to estimate posterior distribution. 
In this survey, we discuss several approximation techniques including variational inference (VI), Monte Carlo dropout (MCD), Markov chain Monte Carlo (MCMC) and Laplace approximation.
A detailed review of each category is provided in the following subsections.\par


\subsubsection{Variational inference (VI)}
\label{sec:sec:vi}
VI~\cite{hinton1993keeping} has been successfully applied as an approximation technique to many applications of neural networks.
It uses a pre-specified distribution $q(\theta)$ to infer the posterior distribution $p(\theta|x,y)$.
In other words, VI aims to make $q(\theta)$ to be close to the posterior obtained from the original model through optimizing a set of parameters.
To achieve this, the Kullback-Leibler (KL) divergence~\cite{kullback1951on} can be defined as: 
\begin{align}
\label{eq:KL}
    KL(q||p)= \mathbb {E}_{q} \Bigg[\log \frac{q(\theta)}{p(\theta|x,y)}\Bigg].
\end{align}
But the KL divergence cannot be directly minimized because of the posterior $p(\theta|x,y)$. Instead, the evidence lower bound (ELBO) can be optimized.  
As such, the ELBO for a given prior distribution over the model parameters can be written as: 
\begin{align}\label{eq:elbo}
    L= \mathbb {E}_{q} \Bigg[\log \frac{p(y|x,\theta)}{q(\theta))}\Bigg].
\end{align}
and for the KL divergence: 
\begin{align}
\label{eq:KLelbo}
    KL(q||p)= -L+\log~p(y|x),
\end{align}
holds.

Graves et al.~\cite{graves2011practical} introduced a stochastic variational method to reduce the difficulty in inferring analytical solutions of the original VI. 
They used numerical integration to approximate the expected values. 
Bayes By Backprop (BBB)~\cite{blundell2015weight} is an extended version of the stochastic variational method~\cite{graves2011practical} to non-Gaussian priors. 
Specifically, BBB uses an unbiased estimate of gradients to learn a distribution over the network's weights.  
Kingma et al.~\cite{kingma2013auto,kingma2015variational} reduced the variance
of the stochastic gradients by introducing a reparameterization strategy. This strategy can approximate posterior inference in models with continuous latent variables. 
Rezende and Mohamed~\cite{rezende2015variational} used normalizing flow to construct distributions for approximation. 
This technique applies a sequence of invertible transformations to transfer a simple density to a more complex one.
Zeng et al.~\cite{zeng2018relevance} estimated the epistemic uncertainty in an active learning framework using Bayesian convolutional neural networks (CNNs) with Gaussian approximate VI.
{\color{black}They showed that using a few Bayesian layers close to the output layer of CNN can estimate a similar level of uncertainty as with that of the original Bayesian CNNs. }

Zhang et al.~\cite{zhang2018noisy} showed that natural gradient descent~\cite{amari1997neural} with adaptive weight noise can be fitted as a variational posterior to maximize the ELBO.
Later, Osawa et al.~\cite{osawa2019practical} trained deep networks using a natural gradient VI, namely variational online Gauss-Newton (VOGN)~\cite{khan2018fast}, and obtained similar results to that of Adam optimizer by using strategies such as batch normalization and data augmentation.   
The \textit{stochastic low-rank approximate natural-gradient} (SLANG)~\cite{mishkin2018slang} is a variant of VI methods that use a structure based on diagonal plus low-rank to compute the Gaussian approximations. In addition, SLANG uses the back-propagated gradients of the network log-likelihood to build a covariance, which enables the model a faster estimation than mean-field methods.
Heo et al.~\cite{heo2018uncertainty} proposed \textit{uncertainty-aware attention}, which uses VI with dropout sampling, to compute the epistemic uncertainty together with the heteroscedastic uncertainty in predicting time-series data.


\subsubsection{Monte Carlo dropout (MCD)}
\label{sec:sec:mmcd}
Monte Carlo (MC)~\cite{neal1993bayesian} is another approximation technique that has been widely used for estimating the posterior distribution, but it is computationally expensive and slow. 
Gal devised MC dropout (MCD)~\cite{gal2016dropout,gal2016uncertainty} to alleviate these issues.
MCD integrates dropout~\cite{srivastava2014dropout}, which is an effective technique for tackling overfitting problems in deep models, as a regularization term to estimate the prediction uncertainty. 
During learning, dropout randomly (with a certain probability $p$) drops some model units to avoid excessive co-tuning.
MCD uses the mean of $N$ models, $f_{\theta_1}, ..., f_{\theta_N}$, parametrized by $\theta_1, ..., \theta_N$ to approximate outcome based on the posterior estimation of the weights as follows~\cite{gal2016dropout}:
\begin{align}\label{eq:MCD}
    y^* \approx \frac{1}{N} \sum_{i=1}^N y^*_i=\frac{1}{N} \sum_{i=1}^N f_{\theta_i}(x^*).
\end{align}

Later, Gal et al.~\cite{gal2017concrete} introduced a new variant of dropout, called concrete dropout, which uses gradient methods instead of grid search to tune the dropout probability. 
This leads to a calibrated uncertainty estimate in large models. 
Mokhoti and Gal~\cite{mukhoti2018evaluating} integrated MCD and concrete dropout as inference techniques into the \textit{DeepLab-v3+}~\cite{chen2018encoder} structure sense segmentation. 
In addition, they introduced a new metric, namely \textit{mutual information}, to estimate epistemic uncertainty by computing the mutual information between a predictive distribution and posterior over the model weights.
The single-shot MCD~\cite{brach2020single} analytically approximates the expected value and variance of the MCD for each layer of the fully connected network. This model requires less computational time as compared with that of the MCD.  
Study~\cite{kennamer2019empirical} conducted an empirical study to show how epistemic uncertainty is affected when the observing condition is changed using MCD.\par

Since the proposal of MCD, many scholars have applied it to estimate epistemic uncertainty. 
For example, Abdar et al.~\cite{abdar2021uncertainty,abdar2021barf} applied MCD to tackle uncertainty during skin cancer image classification.
Studies~\cite{wang2019aleatoric} and~\cite{nair2020exploring} integrated MCD into CNN to estimate the epistemic uncertainty for segmentation and lesion detection in medical images.
Loquercio et al.~\cite{Loquercio2019general} computed the epistemic uncertainty in robotics by combining Bayesian belief networks with MCD. 
Bertoni et al.~\cite{bertoni2019monoloco} estimated the epistemic uncertainty for Monocular 3D Pedestrian Localization using MCD.
In~\cite{zhu2017deep}, MCD is used to estimate the epistemic uncertainty in an encoder-decoder framework with long-short-term-memory (LSTM) for time series forecasting and anomaly detection using Uber data.
Xiao and Wang~\cite{xiao2019quantifying} utilized MCD to estimate the epistemic uncertainty for natural language processing tasks.\par



\subsubsection{Markov chain Monte Carlo (MCMC)}
\label{sec:sec:mcmc}
MCMC is another popular technique to approximate inference and represents epistemic uncertainty. 
It first samples from arbitrary distributions and then performs a stochastic transition governed by the current state and the desired distribution, e.g., true posterior.  
In other words, MCMC starts with generating samples in an iterative and Markov chain fashion.
Markov chain is a distribution over random variables that undergoes a transition from one state to another one in the space state.
At each iteration, the model selects samples based on pre-specified rules.
This process is iterated $T$ times.
Finally, the desired distribution is approximated using the generated samples.  
It aims to sample for a set of independent observations $\textbf{x} \in {D}$ from the posterior distribution $\theta$~\cite{chen2014stochastic}:
\begin{align}
    p(\theta|D)\propto \exp (-U(\theta)),
\end{align}
where $U$ is the potential energy function defined by: 
\begin{align}
    U=-\sum_{x\in D}log~p(x|\theta)-log~p(\theta).
\end{align}

Hamiltonian (hybrid) MC (HMC)~\cite{neal1993bayesian,duane1987hybrid} is the first one that involves using the MCMC sampling technique for Bayesian neural networks.
It explores the state space based on the Metropolis-Hastings framework instead of a random-walk strategy to sample from $\theta$. 
As such, it introduces a set of auxiliary momentum variables, denoted by $r$, from a Hamiltonian system.  
In order to sample from $p(\theta|D)$, HMC generates samples from a joint distribution of $(\theta, r)$ as follows: 
\begin{align}
    \pi (\theta,r)\propto \exp \Bigg(-U(\theta)-\frac{1}{2}r^T~M^{-1}r\Bigg),
\end{align}
where $M$, which is a mass matrix and often set to the identity matrix, together with $r$ indicates a \textit{kinetic energy} term. 
The Hamiltonian function in given by: 
\begin{align}
    H(\theta, r)= U(\theta)+\frac{1}{2}r^T~M^{-1}r.
\end{align}

The Hamiltonian dynamics is simulated by HMC to generate samples, as follows: 
\begin{align}
		        \left\{
                \begin{array}{ll}
                  d\theta = M^{-1}r~dt\\
                  dr=-\bigtriangledown U(\theta)~dt\\
                \end{array}.
              \right.
\end{align}


Despite the success of HMC, it requires processing all data samples at each iteration, which is computationally expensive specifically for large data sets.
To alleviate this issue, many algorithms have attempted to use a mini-batch strategy.   
In this regard, Welling and Teh~\cite{welling2011bayesian} proposed stochastic gradient descent (SGD) HMC method that combines SGD with first-order Langevin dynamics. 
{\color{black}Later, Chen et al.~\cite{chen2014stochastic} proved that using second-order Langevin dynamics can explore the space of solutions and provide good generalization.}
In addition, they added friction into the SGD-HMC to update momentum and evaluated the impact of the noisy gradient. 
Teye et al.~\cite{teye2018bayesian} showed that training deep models with batch normalization is equal to that of estimating the inference in Bayesian networks.
Chandra et al.~\cite{chandra2021bayesian} proposed Bayesian graph deep learning techniques that use MCMC samples with Langevin-gradient.
Mandt et al.~\cite{mandt2017stochastic} used SGD with a constant learning rate (constant SGD) to simulate the Markov chain with a stationary distribution and showed that constant SGD can approximate the posterior inference.
Cyclical stochastic gradient MCMC (SG-MCMC)~\cite{zhang2019cyclical} used a cyclical stepsize schedule to better approximate posterior distributions. 
However, using a mini-batch strategy, that employs a small set of samples at each iteration, adds noise to the network and increases its uncertainty.  
To alleviate this, Luo et al.~\cite{luo2018thermostat} used Nos{\'e}-Hoover thermostats~\cite{hoover1985canonical} to deal with the generated noise.
The resulting method is called \textit{thermostat-assisted continuously tempered HMC}.

Maddox et al.~\cite{maddox2019simple} introduced stochastic weight averaging Gaussian (SWA-Gaussian) to represent uncertainty and calibrate deep models.
Specifically, SWA-Gaussian uses SWA~\cite{izmailov2019averaging} to compute the mean of SGD iterates with a high constant learning rate in order to improve the generalization in deep models.
In addition, the Gaussian posterior approximation over the model weights is approximated by using mean SWA and computing a low-rank plus diagonal approximation to the covariance of the iterates.
Garg and Awate~\cite{garg2018perfect} proposed a perfect/exact MCMC for generic Markov random fields to compute the uncertainty in multi-label segmentation. 
Specifically, they combined two schemes, namely coupling from the past~\cite{gary1996exact} and bounding-chain~\cite{mark2004perfect}, to propose perfect-sample label images. 
Hernández et al.~\cite{hernandez2020improving} combined dropout and HMC to improve the predictive uncertainty in classification problems. 
Akkoyun et al.~\cite{akkoyun2020predicting} applied MCMC into a Bayesian framework to predict maximum aneurysm diameter. 
In addition, Cai et al.~\cite{cai2018uncertainty} developed proximal MCMC techniques to estimate uncertainty in radio interferometric imaging. 

\subsubsection{Laplace approximation}
\label{sec:sec:lap}
As another powerful approximating method, Laplace approximation tackles the problem of representing a complex posterior over the parameters of neural networks by assuming it as a Gaussian distribution~\cite{mackay1992practical}. Different from variational approximation methods, Laplace approximation is a local approximation technique that pays more attention to the trend around the mode of the posterior distribution. As described in~\cite{Hong2019masterthesis}, the expectation $\mu$ of Gaussian distribution $q(\theta)$ is the extreme point $\theta^*$ of posterior distribution $p(\theta|D)$. Thus, $\mu$ is determined by the first derivative of $p(D,\theta)$ which meet the condition $p(\theta|D)\propto p(D,\theta)$, while the covariance matrix $\mathbf{\Sigma}$ is obtained by the second-order Taylor expansion of $\ln{p(D,\theta)}$ centering on $\theta^*$:
\begin{align}
        \ln{p(D,\theta)} \approx \ln{p(D,\theta^*)-\frac{1}{2}(\theta-\theta^*)^\mathsf{T}\mathbf{H}(\theta-\theta^*)},
\end{align}
where $\mathbf{H}$ is Hessian matrix which defined as:
\begin{align}
        \mathbf{H}=-\frac{\partial^2\ln{p(D,\theta)}}{\partial \theta^2}\bigg|_{\theta = \theta^*}.
\end{align}

Then, the posterior $p(\theta|D)$ is approximated as Gaussian $q(\theta)$ with covariance matrix $\mathbf{\Sigma}=\mathbf{H}^{-1}$:
\begin{align}
        p(\theta|D) \approx q(\theta) \sim N(\theta|\theta^*,\mathbf{H}^{-1}).
\end{align}

Unfortunately, it is infeasible to compute the Hessian matrix for deep neural networks with a significant number of parameters. {\color{black} Relatively, constructing a diagonal matrix in curvature approximating for 
a neural network is more calculable and efficient.} Kirkpatrick et al.~\cite{kirkpatrick2017overcoming} used diagonal Laplace approximations to enhance the capability of deep neural networks for sequentially learning tasks by preserving the weights important for previous tasks. Subsequently, Ritter et al.~\cite{ritter2018scalable} first pointed out the limitation of diagonal approximation when some weights exhibit high covariance, then suggested the effectiveness of Kronecker Factorization for acquiring covariance in Laplace approximation and successfully applied in learning online scenarios~\cite{ritter2018online}.


More recently, Lee et al.~\cite{lee2020estimating} developed a sparsification technique using a low-rank approximation to demonstrate the effectiveness of scaling Laplace approximation to large-sized data sets (e.g., ImageNet) and architectures. Schillings et al.~\cite{schillings2020convergence} discussed the convergence of Laplace approximation in Hellinger distance. Margossian et al.~\cite{margossian2020hamiltonian} derived an adjoint method to promote the computation of Monte Carlo with an embedded Laplace approximation in order to marginalize out weights. Daxberger et al.~\cite{daxberger2021bayesian} obtained posteriors by performing inference over a small subset of model weights and outlined the procedure for scaling the linearized Laplace approximation to large neural network models within the framework of subnetwork inference.
A new idea, L2M~\cite{perone2021l2m}, estimated uncertainty by expanding Laplace approximation with gradient raw second moment estimation in optimizers.






\subsection{Ensemble methods}
\label{sec:sec:ensemble}





Ensemble learning is generally aimed at training multiple models that are combined to make a final prediction. In a regression context, simple averaging is a commonly used way of combining multiple models $f_i: X\to Y$ for $i=1, \ldots, M$ in an ensemble $f_{ens}: X\to Y$, as illustrated below: 

\begin{equation}\label{simple averaging}
    f_{ens}(x)= \frac{1}{M}\sum\limits_{i=1}^{M}f_i(x).
\end{equation}

In a classification context, majority voting is a commonly used rule of fusing multiple classifiers to finally output a class $c= 1, 2, \ldots, k$, as illustrated below:

\begin{equation}\label{majority vote}
    f_{ens}(x)= \mathop{\arg\max}\limits_{c} \sum\limits_{i=1}^{M} I(f_i(x)=c).
\end{equation}

In general, it often appears that the outputs of multiple models in an ensemble are different, where the variance of the outputs is considered as an indicator of epistemic uncertainty in a prediction~\citep{hullermeier2021aleatoric,Lakshminarayanan:Ensemble,Tran:Diversity}. On the other hand, the epistemic uncertainty is also referred to as ensemble ambiguity (diversity), which is viewed as a key factor of successful ensemble learning~\cite{Kuncheva:Diversity,Cavalcanti:Diversity}. {\color{black}In this section, we will introduce the general concept of ensemble diversity and analyze 
the importance of the diversity in terms of improving the ensemble performance.} Moreover, we will provide a review of those existing methods of diversity quantification and creation.

\subsubsection{General Concept of Ensemble Diversity}
\label{sec:sec:sec:concept}

    
    

Ensemble diversity is generally related to the generalization error of an ensemble. In a regression context, the generalization error can be decomposed through two well-known schemes, namely, ambiguity decomposition~\cite{Krogh:Diversity} and bias-variance decomposition~\cite{Geman:Bias}.

In terms of ambiguity decomposition, given a weighted averaging ensemble (illustrated in Eq.~\eqref{weighted averaging}), the ensemble error $(f_{ens}-y)^2$ can be decomposed into two terms, i.e., the average error of the based models $\frac{1}{M}\sum\limits_{i=1}^{M}w_i(f_i-y)^2$ and the ensemble ambiguity (diversity) $\frac{1}{M}\sum\limits_{i=1}^{M}w_i(f_i-f_{ens})^2$, as illustrated in Eq.~\eqref{ensemble ambituity}.

\begin{equation}\label{weighted averaging}
    f_{ens}(x)= \sum\limits_{i=1}^{M}w_if_i(x),
\end{equation}
where $w_i$ is the weight of each base model $f_i$ with the constraints $0 \leq w_i \leq 1$ and $\sum\limits_{i=1}^{M}w_i=1$, i.e., $f_{ens}$ is essentially a convex combination of the $M$ base models.

\begin{equation}\label{ensemble ambituity}
    (f_{ens}-y)^2= \frac{1}{M}\sum\limits_{i=1}^{M}w_i(f_i-y)^2-
\frac{1}{M}\sum\limits_{i=1}^{M}w_i(f_i-f_{ens})^2.
\end{equation}

According to Eq.~\eqref{ensemble ambituity}, it is straightforward to derive that the ensemble error is guaranteed to be less than or equal to the average error of the base models, i.e., the higher the diversity among the base models is created, the larger the error reduction would be achieved. However, the increase of the diversity may also cause the increase of the average error of the base models~\cite{Brown:Ensemble}, so it is necessary to get the reasonable trade-off between the diversity and the average error.

As discussed in~\cite{Brown:Ensemble}, the ambiguity decomposition does not take into account the possible changes of the training data distribution or the initialized weights distribution. However, it is essential to measure effectively the expected error on unseen data given a specific distribution of training data or initialized weights. {\color{black}From this point of view, the bias-variance decomposition scheme is considered as a useful tool for analyzing 
the generalization error of an ensemble, given that this scheme exactly takes into account the above mentioned changes of distributions.}

The general formulation of the bias-variance decomposition is shown in Fig.~\ref{Fig:dec} in Section~\ref{sec:intro}. Based on this formulation, three concepts have been defined in \cite{Brown:Ensemble}, namely, the averaged bias, the averaged variance and the averaged co-variance of the $M$ base models, as illustrated below:

\begin{equation}\label{averaged bias}
    \overline{bias}= \frac{1}{M} \sum\limits_{i=1}^{M}(E\{f_i\}-y),
\end{equation}

\begin{equation}\label{averaged variance}
    \overline{var}= \frac{1}{M}\sum\limits_{i=1}^{M}E\{(f_i-E\{f_i\})^2\},
\end{equation}

\begin{equation}\label{averaged co-variance}
    \overline{covar}=\frac{1}{M(M-1)}\sum\limits_{i=1}^{M}\sum\limits_{j\neq i}E\{(f_i-E\{f_i\})(f_j-E\{f_j\})\}.
\end{equation}

According to Eq.~\eqref{averaged bias}-\eqref{averaged co-variance}, we can obtain the bias-variance-co-variance decomposition of the mean square error of an ensemble $f_{ens}$, as shown below:

\begin{equation}\label{bias-variance-co-variance decomposition}
E\{(f_{ens}-y)^2\}=\overline{bias}^2+\frac{1}{M}\overline{var}+(1-\frac{1}{M})\overline{covar}.
\end{equation}

Eq.~\eqref{bias-variance-co-variance decomposition} indicates that the mean square error of an ensemble $f_{ens}$ generally depends on the correlation between those base models, where the correlation is quantified through the third term (the averaged co-variance of the base models). Therefore, it is expected to decrease the co-variance, without affecting the bias and variance~\cite{Brown:Ensemble}.

Moreover, the connection between the ambiguity decomposition and the bias-variance-co-variance decomposition was disclosed in~\cite{Brown:Ensemble}. In particular, according to Eq.~\eqref{ensemble ambituity}, the ensemble error $(f_{ens}-y)^2$ can be decomposed into the average error of $M$ base models $\frac{1}{M}\sum\limits_{i=1}^{M}w_i(f_i-y)^2$ and the ensemble ambiguity $\frac{1}{M}\sum\limits_{i=1}^{M}w_i(f_i-f_{ens})^2$. While assuming that the base models are equally weighted for simplicity, the right hand side of Eq.~\eqref{ensemble ambituity} can be substituted into the left hand side of Eq.~\eqref{bias-variance-co-variance decomposition} to obtain a new formulation as below:

\begin{equation}\label{connection}
   E\{\frac{1}{M}\sum\limits_{i=1}^{M}(f_i-y)^2-
\frac{1}{M}\sum\limits_{i=1}^{M}(f_i-f_{ens})^2\}=\overline{bias}^2+\frac{1}{M}\overline{var}+(1-\frac{1}{M})\overline{covar}.
\end{equation}

Based on Eq.~\eqref{connection}, the following formulations can be obtained after some derivations~\cite{Brown:Ensemble}:

\begin{equation}\label{ambiguity-corvariance}
    \begin{split}
       E\{\frac{1}{M}\sum\limits_{i=1}^{M}(f_i-f_{ens})^2\} = \frac{1}{M}\sum\limits_{i=1}^{M}E\{(f_i-E\{f_i\})^2\}- E\{(f_{ens}-E\{f_{ens}\})^2\}\\
       =\overline{var}-var(f_{ens})=\overline{var}-\frac{1}{M}\overline{var}-(1-\frac{1}{M})\overline{covar}
    \end{split}
\end{equation}

\begin{equation}\label{error-bias-variance}
    E\{\frac{1}{M}\sum\limits_{i=1}^{M}(f_i-y)^2\}= \frac{1}{M} \sum\limits_{i=1}^{M}(E\{f_i\}-y)^2+\frac{1}{M}\sum\limits_{i=1}^{M}E\{(f_i-E\{f_i\})^2\}
    =\overline{bias}^2+\overline{var}
\end{equation}

It can be seen from Eq.~\eqref{ambiguity-corvariance}-\eqref{error-bias-variance} that the variance term $\overline{var}$ relates to both the average error of base models and the ensemble ambiguity, so the subtraction of Eq.~\eqref{ambiguity-corvariance} from Eq.~\eqref{error-bias-variance} gets Eq.~\eqref{bias-variance-co-variance decomposition} back and cancels out the variance term $\overline{var}$ (not $\frac{1}{M}\overline{var}$). Moreover, the fact that the variance term $\overline{var}$ appears in both Eq.~\eqref{ambiguity-corvariance} and Eq.~\eqref{error-bias-variance} indicates that it is generally difficult to simply maximize the ensemble ambiguity without affecting the bias term $\overline{bias}$~\cite{Brown:Ensemble,Zhou:Ensemble}.

The above two error decomposition schemes were generally designed for regression problems and can not be directly applied to classification tasks. In terms of the ambiguity decomposition, Eq.~\eqref{ensemble ambituity} derived in a regression setting can be transformed into Eq.~\eqref{KLD}~\citep{Brown:Diversity} to suit a classification task, while assuming that the $M$ base classifiers are fused by combining the class probability values estimated by these classifiers through using the product rule~\citep{Zhou:Ensemble}. In particular, the KL divergence $D_{KL}(y||f_{ens})$ of the ensemble $f_{ens}$ from the target distribution $y$ of class probability is defined as the ensemble error, which can be decomposed into two terms, namely, the average KL divergence $\frac{1}{M}\sum\limits_{i=1}^{M}D_{KL}(y||f_i)$ of the class probability estimates of base classifiers from the target distribution $y$ and the ambiguity $\frac{1}{M}\sum\limits_{i=1}^{M}D_{KL}(f_{ens}||f_i)$ of the ensemble $f_{ens}$.

\begin{equation}\label{KLD}
    D_{KL}(y||f_{ens})=\frac{1}{M}\sum\limits_{i=1}^{M}D_{KL}(y||f_i)-\frac{1}{M}\sum\limits_{i=1}^{M}D_{KL}(f_{ens}||f_i).
\end{equation}

According to Eq.~\eqref{KLD}, the KL divergence $D_{KL}(y||f_{ens})$ of the ensemble $f_{ens}$ from the target distribution $y$ is guaranteed to be less than or equal to the average KL divergence $\frac{1}{M}\sum\limits_{i=1}^{M}D_{KL}(y||f_i)$ of the class probability estimates of the $M$ base classifiers, i.e., the higher the ambiguity $\frac{1}{M}\sum\limits_{i=1}^{T}D_{KL}(f_{ens}||f_i)$ of ensemble $f_{ens}$ is created, the higher the performance improvement would be achieved.

However, the combination of multiple classifiers can be achieved in various ways, e.g., majority vote and average of class probability distributions, which indicates that Eq.~\eqref{KLD} does not provide a general formulation of ensemble diversity for classification tasks.

In terms of the bias-variance decomposition, as discussed in~\cite{Zhou:Ensemble}, Eq.~\eqref{bias-variance-co-variance decomposition} was derived only for regression problems and similar results cannot be obtained for classification tasks. Therefore, Eq.~\eqref{bias-variance-co-variance decomposition} can not be used as a general formulation of ensemble diversity either. Those methods of diversity quantification in a classification context will be introduced in Section~\ref{sec:sec:sec:review}. Moreover, there is not yet a formally accepted definition of the diversity term~\cite{Zhou:Ensemble,Zhou:DF,Kuncheva:Diversity,Cavalcanti:Diversity}, so existing methods of diversity creation 
were designed heuristically using different definitions and the methods will be reviewed in Section~\ref{sec:sec:sec:review-div}.

\subsubsection{Review of Methods of Diversity Quantification}
\label{sec:sec:sec:review}



In a classification context, if the ensemble prediction is made by averaging the class probabilities estimated by $M$ classifiers, then the ensemble diversity can be measured based on Tumer and Ghosh's framework~\cite{Tumer:Diversity,Tumer:Ensemble}. In particular, suppose that each class $c$ has a true posterior probability $P_d(c|x)$ and another posterior probability $P_{e_i}(c|x)$ estimated by a classifier $f_i$, given a one-dimensional feature vector $x$. In this context, the classification error can be decomposed into the Bayes error and the added error, where the Bayes error is irreducible and the added error $\eta_{e_i}(c|x)$ results from the incorrect estimation of the class posterior probability as illustrated below:

\begin{equation}\label{estimation error}
    P_{e_i}(c|x)= P_d(c|x)+ \eta_{e_i}(c|x).
\end{equation}

If it is assumed that the errors of posterior probability estimation on two classes $a$ and $b$ are independent and identically distributed random variables~\citep{Tumer:Ensemble} with zero mean and variance $\sigma_{\eta_i}^{2}$, then the expected added error of classifier $f_i$ in distinguishing the two classes can be defined as shown below:

\begin{equation}\label{added error}
    E_{add, i}= \frac{\sigma_{\eta_i}^{2}}{P'_{d}(a|x)-P'_{d}(b|x)},
\end{equation}

\noindent where $P'_{d}(a|x)$ and $P'_{d}(b|x)$ are the derivatives of the true posterior probabilities of classes $a$ and $b$. In the case of combing the posterior probabilities estimated by $M$ classifiers, the expected added error can be measured in the way as illustrated below~\cite{Tumer:Ensemble}:

\begin{equation}\label{ensemble error}
    E_{add}^{ensemble}= E_{add}(\frac{1+\delta(M-1)}{M}).
\end{equation}

In Eq.~\eqref{ensemble error}, $\delta$ is a correlation coefficient used for measuring the correlation among the estimation errors made by $M$ base classifiers for each class and is thus a way of quantifying the ensemble diversity. If the estimation errors of the $M$ classifiers are independent, i.e., $\delta=0$, then the expected added error of the ensemble would be $\frac{1}{M}$ as same as the added error of each of the $M$ base classifiers (that are assumed to have the same estimation error). However, if the estimation errors of the $M$ base classifiers are perfectly correlated, i.e., $\delta=1$, then the expected added error of the ensemble would be the same as the added error of each base classifier. Moreover, if the estimation errors of the $M$ base classifiers are negatively correlated, i.e., $\delta<0$, then the expected added error of the ensemble can be reduced even more in comparison with the amount of the error reduction in the case of $\delta=0$~\cite{Kuncheva:Diversity}.

In addition to the average rule of fusion, i.e. averaging the class probabilities estimated by multiple classifiers, majority voting is also a popular rule of combining classifiers. Since the outputs of the $M$ classifiers in a majority vote ensemble are not numeric, the correlation coefficient $\delta$ used in Eq.~\eqref{ensemble error} can not be applied directly. Instead, some researchers have tried to define the classification error diversity qualitatively. For example, a scheme has been suggested in~\cite{1998Combining} to classify error diversity into four levels as follows:

\begin{itemize}
    \item Level 1: At most one of the base classifiers in an ensemble makes incorrect classification for each instance.
    \item Level 2: The majority of the base classifiers in an ensemble make correct classification for each instance.
    \item Level 3: At least one of the base classifiers in an ensemble make correct classification for each instance.
    \item Level 4: All of the base classifiers in an ensemble make incorrect classification for each instance.
\end{itemize}

Furthermore, a decomposition of the majority vote error $E_{maj}$ into the average error $E_{avg}$ of the base classifiers, good and bad diversity was introduced in~\cite{Brown:Diversity}, where good diversity has a positive impact on the error reduction and bad diversity results in a negative impact. In this context, Level 1 and Level 2 diversity would be classified as good ones whereas Level 3 and Level 4 diversity would be classified as bad ones. In the setting of binary classification, i.e., $y\in\{+1, -1\}$, the majority vote error decomposition is shown as below:
\begin{equation}\label{majority vote error}
    E_{maj}= \sum\limits_{x} E_{avg}(x)-\sum\limits_{x} y(x)\bar{f}(x)\frac{1}{M}\sum\limits_{i=1}^{M}Dis_{i}(x),
\end{equation}
where the disagreement $Dis_{i}$ between a base classifier $f_{i}$ and the ensemble $\bar{f}$ is measured using:

\begin{equation}\label{disagreement}
    Dis_{i}(x)= \frac{1}{2}(1-f_i(x)\bar{f}(x)).
\end{equation}

In Eq.~\eqref{majority vote error}, the sign of $y(x)\bar{f}(x)$ essentially reflects whether the ensemble classification is correct or not, i.e., $y(x)\bar{f}(x)=+1$ represents correct classification and $y(x)\bar{f}(x)=-1$ indicates incorrect classification. Therefore, Eq.~\eqref{majority vote error} can be rewritten as below:

\begin{equation}\label{good and bad diversity}
     E_{maj}=\sum\limits_{x} E_{avg}(x)-\sum\limits_{x+} \frac{1}{M}\sum\limits_{i=1}^{M}Dis_{i}(x)+\sum\limits_{x-} \frac{1}{M}\sum\limits_{i=1}^{M}Dis_{i}(x).
\end{equation}

\noindent where the second term $\sum\limits_{x+} \frac{1}{M}\sum\limits_{i=1}^{M}Dis_{i}(x)$ denotes good diversity and the third term $\sum\limits_{x-} \frac{1}{M}\sum\limits_{i=1}^{M}Dis_{i}(x)$ denotes bad diversity.

\begin{table}[ht!]
    \centering
    \caption{Contingency table for classifiers $f_i$ and $f_j$}
    \begin{tabular}{c|c|c}
    \hline\noalign{\smallskip}
         & $f_i$ correct(1) & $f_i$ incorrect(0)\\
       \noalign{\smallskip}\hline\noalign{\smallskip}
       $f_j$ correct(1) & $N^{11}$ & $N^{10}$\\
       $f_j$ incorrect(0) & $N^{01}$ & $N^{00}$\\
       \hline
    \end{tabular}
    \label{contingency table}
\end{table}

Those representative methods of the diversity quantification have been analysed in~\cite{Kuncheva:Diversity,Zhou:Ensemble}, which are put into two main categories, namely, pairwise measures and non-pairwise measures. {\color{black} In particular, pairwise measures are generally designed by involving calculation based on a contingency table shown in Table~\ref{contingency table} for a pair of classifiers $f_i$ and $f_j$, whereas non-pairwise measures generally involve counting the number $l(x)$ of classifiers that correctly classify sample $x$ and calculating the relevant probability $P(l(x))$.} 

Those popularly used pairwise measures include the Q-statistic $Q$, the correlation coefficient $\rho$, the disagreement measure $dis$ and the double-fault measure $DF$. Representative non-pairwise measures of diversity include entropy $ENT$, Kohavi-Wolpert variance $KW$, measurement of inter-rater agreement $IRA$, difficulty measure $DM$, generalized diversity $GD$ and coincident failure diversity $CFD$. More details of these diversity measures are summarised in Table~\ref{summary}.

\begin{table}[tb!]
    \centering
    \caption{Summary of Diversity Measures}
    \resizebox{\textwidth}{!}{
    \begin{tabular}{l|l|l|l|l|l}
    \hline\noalign{\smallskip}
      Name   &  Symbol & $\uparrow$/$\downarrow$ & P & Equation & Range\\
      \hline
      Q-statistic & $Q$ & ($\downarrow$) & Y & $Q_{i,j}= \frac{N^{11}N^{00}-N^{10}N^{01}}{N^{11}N^{00}+N^{10}N^{01}}$ & $[-1, 1]$\\
      correlation coefficient & $\rho$ & ($\downarrow$) & Y & $\rho_{i,j}=\frac{N^{11}N^{00}-N^{10}N^{01}}{\sqrt{(N^{11}+N^{10})(N^{01}+N^{00})(N^{11}+N^{01})(N^{10}+N^{00})}}$ & $[-1, 1]$\\
      disagreement measure & $dis$ & ($\uparrow$) & Y & $ dis_{i,j}=\frac{N_{01}+N_{10}}{N^{11}+N^{10}+N^{01}+N^{00}}$ & $[0, 1]$ \\
      double-fault measure & $DF$ & ($\downarrow$) & Y & $DF_{i,j}= \frac{N_{00}}{N^{11}+N^{10}+N^{01}+N^{00}}$ & $[0, 1]$\\
      entropy & $ENT$ & ($\uparrow$) & N & $ENT= \frac{1}{|DS|}\sum\limits_{x\in DS} \frac{1}{M-\lfloor M/2 \rfloor}min\{l(x), M-l(x)\}$ & $[0, 1]$\\
      Kohavi-Wolpert variance & $KW$ & ($\uparrow$) & N & $KW= \frac{1}{|DS|M^2}\sum\limits_{x\in DS} l(x)(M-l(x))$ & $[0, 1]$\\
      inter-rater agreement & $IRA$ & ($\downarrow$) & N & $IRA= 1-\frac{\frac{1}{M}\sum\limits_{x\in DS}l(x)(M-l(x))}{|DS|(M-1)\bar{p}(1-\bar{p})}$ & $[0, 1]$\\
      difficulty measure & $DM$ & ($\downarrow$) & N & $DM= Var(\frac{l(x)}{M})$ & $[0, 1]$ \\
      generalized diversity & $GD$ & ($\uparrow$) & N & $GD= 1-\frac{\sum\limits_{l(x)=1}^{M}\frac{M-l(x)}{M}\frac{(M-l(x)-1)}{(M-1)}p(l(x))}{\sum\limits_{l(x)=1}^{M}\frac{M-l(x)}{M}p(l(x))}$ & $[0, 1]$\\
      coincident failure diversity & $CFD$ & ($\uparrow$) & N & $CFD= \left\{
    \begin{array}{lr}
    0, p(0)=1.0;\\
    \frac{1}{1-p(0)}\sum\limits_{l(x)=1}^{M}\frac{l(x)}{M-1}p(l(x)), p(0)<1
    \end{array}\right\}$ & $[0, 1]$\\
    \hline
    \end{tabular}}
    \label{summary}
\end{table}

In terms of Q-statistics, the value of $Q_{i,j}$ is ranged in $[-1, 1]$ and is expected to be 0 for two statistically independent classifiers $f_i$ and $f_j$. While the two classifiers provide the same outputs correctly, the value of $Q_{i,j}$ tends to be positive~\cite{Kuncheva:Diversity}. In contrast, if the two classifiers make incorrect classifications on different instances, the value of $Q_{i,j}$ would be rendered negative~\cite{Kuncheva:Diversity}. Therefore, it can be concluded that the lower the value of $Q_{i,j}$ resulting from a pair of classifiers $f_i$ and $f_j$, the higher the diversity between $f_i$ and $f_j$ is created.

The value of the correlation coefficient $\rho_{i,j}$ is also ranged in $[-1, 1]$. According to formulations of $Q_{i,j}$ and $\rho_{i,j}$, it is straightforward to identify that the values of $Q_{i,j}$ and $\rho_{i,j}$ always obtain the same sign but $|\rho_{i,j}|\geq|Q_{i,j}|$~\cite{Kuncheva:Diversity,Zhou:Ensemble}.

According to the formulations of the disagreement measure $dis_{i,j}$ and the double-fault measure $DF_{i,j}$, it is easy to see that both $dis_{i,j}$ and $DF_{i,j}$ are ranged in $[0, 1]$. However, $dis_{i,j}$ and $DF_{i,j}$ aim at measure of diversity in different levels (according to the scheme suggested in~\cite{1998Combining} for classifying the diversity into four different levels), i.e., $dis_{i,j}$ shows the percentage of samples that are classified incorrectly by only one of the two base classifiers $f_i$ and $f_j$, whereas $DF_{i,j}$ is the proportion of samples that are mis-classified by both classifiers.

While it is needed to measure the diversity among multiple classifiers, the above-introduced pairwise measures can be used by averaging the values obtained for all those pairs of classifiers. For example, the $Q$ statistic can be used for diversity measure through averaging the values of $Q$ obtained for all classifier pairs, as shown below:

\begin{equation}\label{average Q statistics}
    Q_{avg}= \frac{2}{M(M-1)}\sum\limits_{i=1}^{M-1}\sum\limits_{k=i+1}^{M} Q_{i,j}.
\end{equation}

All the non-pairwise measures are ranged in $[0, 1]$. In the formulations of $ENT$, $KW$ and $IRA$, $|DS|$ represents the number of samples in a data set $DS$ and $l(x)$ denotes the number of classifiers that correctly classify sample $x$. In addition, $\bar{p}$ used in the formulation of $IRA$ denotes the average accuracy of base classifiers. 

In terms of the entropy $ENT$, the minimum value is obtained when all the $M$ classifiers provide the same outputs and the maximum value is obtained when $\lfloor M/2 \rfloor$ classifiers consistently classify sample $x$ as one class and the other $M-\lfloor M/2 \rfloor$ classifiers consistently output another class for $x$. As emphasized in \cite{Kuncheva:Diversity}, the Kohavi-Wolpert variance $KW$ is correlated to the average disagreement measure $dis_{avg}$ by a coefficient $\frac{M-1}{2M}$, i.e., $KW= \frac{M-1}{2M}dis_{avg}$. Moreover, the inter-rater agreement $IRA$ is correlated to both $KW$ and $dis_{avg}$~\cite{Kuncheva:Diversity}, i.e., $IRA= 1- \frac{M}{|DS|(M-1)\bar{p}(1-\bar{p})}KW= 1-\frac{1}{2\bar{p}(1-\bar{p})}Dis_{avg}$. In terms of the difficult measure $DM$, it is generally expected that each instance is difficult for some classifiers but is easy for the other classes to encourage the ensemble diversity~\cite{Kuncheva:Diversity,Zhou:Ensemble}. {\color{black} The minimum of $DM$ is obtained while each instance is easy for the majority of the 
base classifiers, and the maximum of $DM$ is obtained while each instance is either easy or difficult for all the base classifiers.} The generalized diversity $GD$ is designed based on the argument that the incorrect output of one classifier is always accompanied by the correct output of another classifier for maximizing the diversity~\cite{1997Software}. Furthermore, the coincident failure diversity $CFD$ is defined as a modification of $GD$~\cite{1997Software}, which expects that each instance can be classified correctly by some of the base classifiers. While each instance is classified incorrectly by at most one base classifier, the maximum of $CFD$ would be reached, i.e., the highest level of diversity is reached according to the scheme suggested in~\cite{1998Combining} for identifying the level of diversity. 

More recently, Yin et al presented a formulation of diversity learning in~\cite{ConvexEnsemble} as shown below:

\begin{equation}\label{diversity learning}
    \min\limits_{w}f_{loss}(w)-\beta f_{diversity}(w) \hspace{1cm} s.t. \hspace{1cm} w \geq 0
\end{equation}

\noindent where the diversity is treated as a regularization term, $\beta$ is used as a control parameter for the diversity regularization and $w$ represents the model parameter. A formulation of sparsity learning was also presented in~\cite{ConvexEnsemble} for the purpose of ensemble pruning.

Based on the work presented in~\cite{ConvexEnsemble}, more studies have been conducted later on by using diversity as a regularization term~\cite{Cavalcanti:Diversity,Ahmed:Diversity,Dai:Diversity,Dvornik:Diversity,Zhang:Diversity,Bian:Diversity,Wu:Diversity}, such that the ensemble accuracy and diversity can be optimized simultaneously. In particular, Cavalcanti et al~\cite{Cavalcanti:Diversity} proposed to combine different pairwise measures of diversity for ensemble pruning, while the genetic algorithm is used to optimize the combined diversity to obtain several candidate ensembles that are evaluated using the validation data for selecting the final ensemble. Ahmed et al~\cite{Ahmed:Diversity} made an empirical investigation on whether combining the ensemble accuracy with several popular diversity measures is a better evaluation function than using only the accuracy in the setting of ensemble pruning. Dai et al pointed out in~\cite{Dai:Diversity} that accuracy and diversity are closely related to each other and should be considered simultaneously for ensemble pruning. Accordingly, they proposed three new measures for ensemble pruning, namely, Simultaneous Diversity \& Accuracy, Diversity-Focused-Two and Accuracy-Reinforcement. Dvornik et al~\cite{Dvornik:Diversity} proposed to encourage the ensemble diversity by enabling the classifiers to output consistently the highest probability for the ground truth class label and to rank inconsistently the other classes by making different classes obtain the second-highest probability (or other lower-ranked probability). Zhang et al~\cite{Zhang:Diversity} proposed to construct a diversified ensemble layer for combining multiple neural networks as individual modules, while the cross entropy loss of each individual module and the diversity among different modules are optimized simultaneously. Bian et al~\cite{Bian:Diversity} formulated the relationship between the diversity and the ensemble performance in the context of the theorem of margin and generalization and proposed two diversity-driven pruning methods to utilize the formulated relationship, leading to the enhancement of diversity and the reduction of the ensemble size without much loss of performance. Wu et al~\cite{Wu:Diversity} revised those representative diversity measures and introduced focal model based measures of diversity for improving further the correlation between the diversity and the prediction accuracy. Overall, all of the above-reviewed works indicate that effective selection and combination of diversity measures would be essential, such that simultaneous optimization of the ensemble accuracy and diversity can be achieved effectively. 

\subsubsection{Review of Methods of Diversity Creation}
\label{sec:sec:sec:review-div}

In general, ensemble diversity can be created using various types of methods. In this section, we provide a detailed review of diversity creation methods that fall in the category of data input manipulation. We also briefly introduce other methods in the following categories: data output manipulation, manipulation of model architectures, differentiation of starting points in hypothesis space and diversification of learning strategies.

In the setting of data input manipulation, some popularly used methods include Bagging~\cite{Breiman1996}, Random Subspace~\cite{HO:RS} and Boosting~\cite{Freund:Boosting}. Bagging involves training $M$ independent classifiers on $M$ different sample sets drawn by random sampling from the original training set with replacement over $M$ iterations. In this setting, the diversity is created heuristically through diversifying the training samples, which contributes to the variance reduction in the context of bias-variance trade-off~\cite{Bauer:Wagging}. Moreover, two variants of Bagging, namely, Dagging~\cite{Ting:Dagging,Ting:Ensemble} and Wagging~\cite{Bauer:Wagging}, were developed by setting different ways of diversifying the training samples. In particular, Dagging involves drawing $M$ equal-sized sample sets by partitioning the original training set into $M$ disjoint training subsets, whereas Wagging involves learning each of the $M$ base classifiers from the entire training set but each training sample is assigned a random weight. 

Random Subspace can be viewed as a way of diversity creation through feature sampling, which involves training $M$ independent classifiers on $M$ different feature subsets produced by random sampling of features from the full feature set without replacement. Therefore, the Random Subspace method aims at creating diversity heuristically through diversifying the features. A variant of Random Subspace, which is referred to as `Attribute Bagging'~\cite{Brylla:RS}, was designed to require a suitable subspace size to be set as a hyper-parameter for drawing $M$ feature subsets. However, in the setting of Random Subspace, all the base classifiers are learned from the entire training set without diversity creation through manipulation of training samples. In order to better enhance the diversity among base classifiers through the combination of Bagging and Random Subspace~\cite{Fawagreh:RF}, the Random Forest method~\cite{Breiman2001} has been developed and used as a powerful decision tree ensemble approach~\cite{Bernard:RF}.

\begin{figure}[tb!]
\centering
\includegraphics[width=1\linewidth]{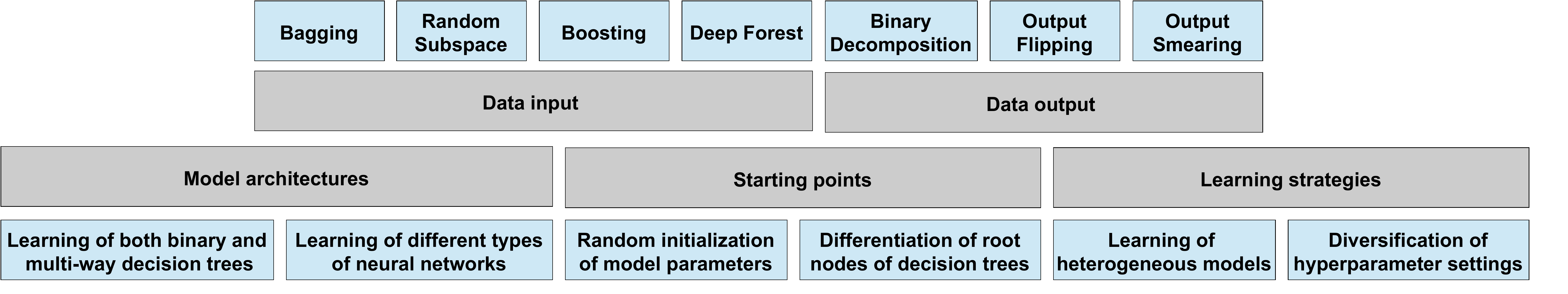}
\caption{ Methods of diversity creation in Ensemble Learning.}
\label{Fig:diversity}
\end{figure}

In contrast to Bagging and Random Subspace that create diversity heuristically leading to independent classifiers, the Boosting approach aims at creating diversity explicitly by training one classifier that aims at correcting the errors resulting from the classifier trained at the previous iteration, i.e., training negatively correlated base classifiers. In particular, two popular methods of Boosting are referred to as `Adaptive Boosting' (AdaBoost)~\cite{Freund1996} and `Gradient Boosting'~\cite{Friedman:GBT}. The former method involves diversity creation through re-weighting samples at each iteration $i$ of learning a base classifier $f_i$. In other words, at the end of each iteration $i$, the weight of each correctly classified sample is decreased and the weight of each misclassified sample is increased, so the learning of classifier $f_{i+1}$ will focus more on those misclassified samples. The re-weighting of each sample $e$ is operated in the way shown below:

\begin{align}\label{sample reweighting}
   \omega_{e}^{i+1} &= \frac{\omega_{e}^{i}exp(-\alpha_i y_{e} f_i(x_e))}{Z_i} \notag\\
   &=\left\{
    \begin{array}{lr}
     \omega_{e}^{i}exp(-\alpha_i), & \text{ if } y_{e} = f_i(x_e)\\
     \omega_{e}^{i}exp(\alpha_i), & \text{ if } y_{e} \neq f_i(x_e)\\
   \end{array}\right\}.\notag\\
\end{align}
    
In Eq.~\eqref{sample reweighting}, $\omega_{e}^{i}$ is the weight of sample $e$ updated at iteration $i$, $Z_i$ is a normalization factor (Eq.\eqref{normalization factor}), $\alpha_i$ is the weight of classifier $f_i$ (Eq.~\eqref{classifier weight}) measured based on the classification error rate $\epsilon_i$ (Eq.~\eqref{error rate}), $x_e$ represents the feature vector of sample $e$ and $y_{e}$ represents the ground truth label of sample $e$.

\begin{equation}\label{normalization factor}
    Z_i= \sum_{e=1}^{N}\omega_{e}^{i}exp(-\alpha_i y_{e} h_i(x_e)),
\end{equation}

\begin{equation}\label{classifier weight}
    \alpha_i= \frac{1}{2}\ln(\frac{1-\epsilon_i}{\epsilon_i}),
\end{equation}

\begin{equation}\label{error rate}
    \epsilon_i= \frac{\sum_{e=1}^{N} \omega_{e}^{i} \cdot I(y_{e} \neq f_i(x_e))}{\sum_{e=1}^{N} \omega_{e}^{i}}.
\end{equation}

In Eq.~\eqref{error rate}, $I(\cdot)$ is an indicator function comparing the ground truth label $y_e$ and the output label $f_i(x_e)$ for each sample $e$.

According to Eq.~\eqref{sample reweighting}-\eqref{error rate}, it can be derived that classifier $f_i$ trained at iteration $i$ must classify correctly as many as possible those samples misclassified by the ensemble $\bar{f}_{i-1}$ of classifiers trained at the previous iterations, in order to reduce the error rate $\epsilon_i$. Also, due to the case that classifier $f_i$ pays less attention to those samples that were classified correctly by the ensemble $\bar{f}_{i-1}$ of classifiers trained at the previous iterations, some of such samples may be misclassified by classifier $f_i$. Therefore, the AdaBoost method is considered to be able to create diversity among classifiers explicitly. 

In contrast to AdaBoost, the Gradient Boosting method is aimed at training a classifier at iteration $i$ to fit the negative gradient (residual) estimated at iteration $i-1$, based on the error rate of the ensemble $\bar{f}_{i-1}$ of the classifiers trained at the previous iterations. Moreover, any differential loss functions can be used in the setting of gradient boosting, which overcomes the limitations of the AdaBoost method in terms of the selection of loss functions. Based on the principle of the Gradient Boosting method, a popular decision tree ensemble method, referred to as Gradient Boosting Decision Trees (GBDT)~\cite{Hastie:Statistic}, has been developed and used in many application areas. In a regression context, as illustrated in Eq.~\eqref{residual}, GBDT is essentially aimed at training a decision tree $f_i$ at each iteration $i$ by optimizing $\theta_i$ to fit the residual $r_{i-1}$ resulting from the tree ensemble $\bar{f}_{i-1}$ obtained at iteration $i-1$ for error reduction, i.e., the better the decision tree $f_i$ fits the residual $r_{i-1}$, the larger the error reduction would be achieved. As pointed out in~\cite{Zhou:Ensemble}, the error reduction is achieved primarily through the bias reduction although the variance reduction can also be achieved.
\begin{equation}\label{residual}
    L(y, \bar{f}_{i-1}(x)+f_{i}(x; \theta_i))= (y-\bar{f}_{i-1}(x)-f_{i}(x; \theta_i))^2=(r_{i-1}-f_{i}(x; \theta_i))^2.
\end{equation}

Based on the Bagging, Random Subspace and Boosting approaches, there have been a variety of decision tree ensemble methods developed by introducing specific ways of diversity creation.  In particular, Dynamic Random Forest~\cite{Bernard:RF} (a variant of Random Forest) involves training $M$ decision trees on $M$ training sets with different weight distributions over those samples, where the weight~$\omega_{e}^{i}$ of each sample $e$ at each iteration~$i$ is set heuristically to be equal to the proportion of base classifiers correctly classifying sample $e$ to the total number of base classifiers obtained so far. Rotation Forest~\cite{Rodriguez:Forest} involves using Principal Component Analysis (PCA)~\cite{Hastie:Statistic} over $M$ iterations to draw $M$ transformed feature sets, such that $M$ diverse decision trees are trained. Furthermore, Rotation Random Forest~\cite{Zhang:RF} was developed as a variant of Rotation Forest, which involves using PCA or Linear Discriminant Analysis (LDA)~\cite{Hastie:Statistic} to transform each feature subset selected randomly for generating each specific node of a decision tree. Extremely Randomized Trees (Extra-Tree)~\cite{Geurts:Forest} involves not only randomly selecting a feature subset for generating each specific node of a decision tree but also randomly selecting a numeric value as the cut-point at the tree node if the split attribute is continuous. Random Feature Weights for decision tree ensemble construction~\cite{Maudes:Forest} was designed to assign each feature a random weight (ranged in [0, 1]) for training a decision tree at each iteration~$i$. In this setting, $M$ different decision trees are trained using $M$ feature sets with different weight distributions over those features. Forest by Penalizing Attributes (Forest PA)~\cite{Adnan:FPA} was designed to assign each attribute $Attr$ a weight heuristically at each iteration $i$, based on the level of the tree (trained at iteration $i-1$) in which the node (corresponding to attribute $Attr$) is located.

In addition to the above introduced methods, it is also a popular strategy of data input manipulation to create diversity through training multiple classifiers on different sets of features extracted in different ways~\cite{Gunes:Fusion,2017Multimodal}. In the era of deep learning, a new type of decision tree ensemble referred to as `Deep Forest'~\cite{Zhou:DF,Zhou:Forest} has become more popular. The pilot study was reported in~\cite{Zhou:DF}, which introduces the gcForest method that aims at producing a cascade of decision forests, i.e., creating an ensemble of ensembles. In particular, the major idea of gcForest is to train a model that involves multiple layers and multiple decision forests (an ensemble of decision tree ensembles) in each layer, where the feature space is dynamically changed every time a new layer is added. In other words, some new features, which are generated as outputs in each layer $lr_{i}$, are used as inputs for the next layer $lr_{i+1}$, where all the original features are kept for each layer. In this context, the ensembles of decision forests in different layers are produced using different sets of features, so those ensembles of decision forests trained in different layers are considered to involve diversity created through diversification of feature sets. Based on gcForest, some variants have been developed later on through setting different strategies of feature space update, such as multi-layered GBDT~\cite{Feng:GBDT}, Deep Extra-Tree~\cite{Berrouachedi:DF}, Deep Multigrained Cascade Forest~\cite{Liu:DF}, Densely Connected Deep Random Forest~\cite{Cao:DF}, Rotation-based deep forest~\cite{Cao:Rotation}, Siamese Deep Forest~\cite{Utkin:DF}.

In terms of data output manipulation, a popular way is to transform a multi-class classification problem into a number of binary classification problems through binary decomposition. Popular strategies of decomposition include one-vs-one (OVO)~\cite{Gutierrez:Ordinal}, one-vs-rest (OVR)~\cite{Wozniak:Ensemble}, many-vs-many (MVM) strategy~\cite{Gutierrez:Ordinal}. In the setting of binary decomposition, an ensemble of binary classifiers is created and the error correcting output codes (ECOC) strategy~\cite{Dietterich:ECOC} is commonly used for fusing the outputs of the binary classifiers to finally classify a new sample. Also, ECOC has shown its effectiveness in improving the diversity between binary classifiers in the setting of end-to-end neural network training~\cite{Song:Diversity}. More recently, N-nary decomposition has been proposed in~\cite{Zhou:Ordinal} as a generalization of binary decomposition. In addition, two other representative ways of output manipulation for diversity enhancement are referred to as `Output Flipping' and `Output Smearing', which have been proposed and experimented in~\cite{Breiman:Flipping}.

In addition to data manipulation, some other ways can also be taken in practice for diversity creation, which include manipulation of model architectures, differentiation of starting points in the hypothesis space and diversification of learning strategies. The manipulation of model architectures can be applied to decision tree learning, leading to a tree ensemble that contains both binary and multi-way trees, e.g., combining a binary tree trained by CART and two multi-way trees trained by ID3 and C4.5~\cite{Bashir:Forest}. Also, in the setting of neural network learning, different types of networks can be produced to form an ensemble through manipulating the network architectures~\cite{Zhang:Diversity}. Differentiation of starting points in the hypothesis space can be applied to neural network learning through random initialization of weights~\cite{Brown:Ensemble} over multiple iterations for training complementary models. In the setting of decision tree learning, starting points in the hypothesis space can be differentiated by selecting different attributes for the root nodes of the trees~\cite{Adnan:CERN}. In addition, diversification of learning strategies can be achieved by training heterogeneous classifiers through using different learning algorithms~\cite{Zhou:DF,Zhou:Forest}, or using different hyper-parameter settings of the same learning algorithm, e.g., combination of various loss functions~\cite{Hajiabadi:Ensemble}.

\section{Applications}
\label{sec:app}

This section discusses the importance of estimating epistemic uncertainty in several popular applications. 
These applications include computer vision and natural language processing (NLP).
In the following subsections, we first review the applications of epistemic uncertainty learning in computer vision, and then explain how epistemic uncertainty learning has applied to NLP.

\subsection{Computer vision}
\label{sec:sec:cv}
{\color{black} In computer vision, uncertainty is taken into account in variety of applications such as image classification~\cite{abdar2021uncertaintyfusenet,senousy2021mcua}, segmentation~\cite{cai2018uncertainty,kwon2020uncertainty}, camera relocalization~\cite{kendall2016modelling}, object detection~\cite{schubert2020metadetect,catak2021prediction,chen2021generating},  image/video retrieval (restoration)~\cite{zhang2019reducing,dorta2018structured}, in the setting of 
Bayesian and ensemble learning.} 
Image classification and segmentation are among the most popular applications of DL models. The former categorize all objects in an image into a single class, while later aims to assign a label to each pixel in a single image in which pixels from a label share specific properties.\par

Both classification and segmentation have been widely used for medical image analysis.
Although the state-of-the-art supervised learning models can produce precise predictions, they are uncertain about the quality of their predictions.  
Since the size and shape of diseases are different, and they locate across the patient's body, it is vital to address uncertainties and make predictions interpretable and reliable. 
Known et al.~\cite{kwon2020uncertainty} proposed an uncertainty estimation method using the Bayesian neural networks for stroke lesion segmentation. 
This method finds a relationship between variance and means of a multi-modal random value. 
Abdar et al.~\cite{abdar2021uncertaintyfusenet} integrated an ensemble MCD into a multi-model learning framework, which receives chest X-ray (CXR) and computed tomography (CT) images as inputs, to estimate uncertainty in identifying COVID19 cases.
Study~\cite{araujo2020drgraduate} proposed an uncertainty-aware framework for grading diabetic retinopathy. This framework built a Gaussian sampling approach based on multiple instance learning strategies to infer the grade of images.\par

Object detection is another popular application of supervised learning models that are being extensively used in autonomous cars.
Any mistake in their predictions may cause catastrophic damages or even fatality; therefore, it is vital to estimate the reliability of their predictions.      
In this regard, prediction surface uncertainty~\cite{catak2021prediction}, denoted as PURE, was proposed to estimate the predictive uncertainty. This model formulates the object detection task as a regression problem to locate objects in a 2D-camera view image and uses MCD to estimate the uncertainty of the model.
Study~\cite{li2021uncertainty} proposed an uncertainty-aware model for the detection of both salient and camouflaged objects. Specifically, the contradicting attributes of these two tasks were modeled using a similarity measure technique. In addition, an adversarial learning model was proposed to compute the network confidence score.\par





As the basis of downstream image classification and segmentation tasks, image restoration is an inverse image degradation process. {Specifically, it processes the degraded image caused 
by the imaging device subject to external interference and restores a high-quality image approximating the original image before being degraded.} {\color{black}In image restoration tasks, the degraded images are samples with high-level aleatoric 
uncertainty. 
Study~\cite{zhang2019reducing} estimated uncertainty resulting from the undersampled source data. It enhanced the quality of reconstructed images by utilizing a specific network branch to study inherent aleatoric 
uncertainty arising from noise data.} While epistemic uncertainty was superb at estimating the reliability of restored images, satisfying the requirements of safety-critical fields, such as magnetic resonance (MR) images reconstruction. As Begoli et al.~\cite{begoli2019need} concluded that understanding prediction system structure and defensibly quantifying uncertainty is significantly beneficial for medical AI applications. 
Tanno et al.~\cite{tanno2017bayesian} combined a 3D subpixel-CNN based framework with Bayesian image quality transfer (IQT)~\cite{tanno2016bayesian} to solve diffusion MRI reconstruction problems. They described intrinsic uncertainty as an irreducible variance of mapping low-resolution(LR) to high-resolution(HR) images and defined the degree of ambiguity in the model parameters as parameter uncertainty captured by variational dropout. Subsequently, Schlemper et al.~\cite{schlemper2018bayesian} introduced MCD into reconstruction networks, demonstrating the competitive performance of quantifying epistemic uncertainty by utilizing Bayesian methods, especially dealing with test samples which out of training data distribution and superior to overparametrised deterministic networks.

Epistemic uncertainty learning techniques have been applied to other image restoration tasks such as denoising~\cite{cheng2019bayesian,serra2017bayesian}, deraining~\cite{chen2021robust}. For instance, Cheng et al.~\cite{cheng2019bayesian} presented an MCMC-based Stochastic gradient Langevin dynamics (SGLD) framework to approximate the posterior distribution to improve performance in image denoising tasks. Serra et al.~\cite{serra2017bayesian} proposed a fast variational inference framework for solving the sparse representation-related problems in image processing and successfully applied it to the denoising problem.


{\color{black} In addition, several studies have addressed the epistemic uncertainty in analyzing 
video streams.} Huang et al.~\cite{huang2018efficient} utilized the similarity of consecutive frames,i.e., temporal property, in videos. They proposed region-based temporal aggregation (RTA) framework, which dramatically speeds up MC-dropout in video segmentation tasks, to estimate uncertainty by calculating the moving average of prediction in consecutive frames to simulate the sampling procedure. Study~\cite{zhao2019generative} is the first learning-based solution for the bronchoscopic localization, which estimates uncertainty utilizing VI to conduct video-CT registration.

\subsection{Natural language processing}
\label{sec:sec:nlp}

In natural language processing tasks, various metrics of uncertainty quantification have been studied~\cite{xiao2019quantifying,Dong:NLP,Shen:Uncertainty,Wang:NLP,He:NLP} in the context of either Bayesian deep learning or ensemble learning. 

{\color{black}In setting of Bayesian deep learning, Xiao and Wang~\cite{xiao2019quantifying} proposed novel methods of quantifying epistemic and aleatoric 
uncertainties in sentiment analysis, named entity recognition and language modeling tasks, and the experimental results show that learning to quantify uncertainty is not only necessary in measuring the prediction confidence but also useful in improving the model performance.} Dong et al.~\cite{Dong:NLP} outlined three major causes of uncertainty and designed various metrics for quantifying these factors and estimating confidence scores that indicate the likelihood of correct predictions made by a model. The experimental results reported in~\cite{Dong:NLP} show that the proposed confidence model outperforms those methods that rely on confidence scores based on posterior probability, and the interpretation of uncertainty is also improved in comparison with simply using attention scores. Wang et al.~\cite{Wang:NLP} proposed to quantify the epistemic uncertainty for measuring the prediction confidence of a neural machine translation model and their experimental results indicate that the performance of machine translation can be improved significantly through uncertainty-based estimation of prediction confidence.  

In the setting of ensemble learning, Shen et al~\cite{Shen:Uncertainty} investigated applying Gaussian processes and random forests for measuring the uncertainty in document quality predictions. The experimental results reported in~\cite{Shen:Uncertainty} indicate that both Gaussian processes and random forests can be used effectively in predicting the quality of Wikipedia articles alongside an estimate of the uncertainty concerning the inconsistent outputs of various models. He et al.~\cite{He:NLP} proposed to improve the confidence of winning score for generating accurate uncertainty score. In particular, a model, which consists of three parts, namely, ``mix-up", ``self-ensembling" and ``distinctiveness score", is proposed in the setting of deep neural networks for reducing the impact of the overconfidence of winning score and also taking into account the impacts of other types of uncertainty. The experimental results reported in~\cite{He:NLP} indicate that accurate scores of uncertainty can be obtained using the proposed model and the performance of text classification can be improved by assigning those uncertain predictions to domain experts.


\section{Discussions and Conclusions}
\label{sec:discuss}


In this survey, we provided a hierarchical categorization of the epistemic (model) uncertainty learning methods, i.e., 
Bayesian and ensemble methods. Bayesian methods formulate epistemic uncertainty as a posterior distribution over the weight parameters. Since these methods need to compute posterior, they cannot perform inference analytically but can be approximated. In this regard, we discuss four widely used approximation techniques, including variational inference (VI), Monte Carlo dropout (MCD), Markov Chain Monte Carlo (MCMC), and Laplace approximation. Each of these techniques has several advantages and disadvantages~\cite{abdar2021review}. Among them, MCD techniques are easy to implement and don't need to change the training process. However, they are not reliable for out-of-distribution samples and require multiple sampling when performing inference. VI techniques benefit from stochastic optimization methods and are suitable for big data sets. However, they are computationally complex.MCMC techniques can approximate exact posterior, but they are very slow and fail to converge. Although the Laplace approximation techniques have a simple procedure, they perform poorly due to ignoring the global properties of the real posterior.\par

In contrast, ensemble methods formulate epistemic uncertainty as the variance of the outputs of base models. The epistemic uncertainty is also referred to as ensemble diversity, which is considered as a key factor of successful ensemble learning. In particular, existing works have illustrated mathematically how the ensemble diversity impacts the generalization performance in the context of bias-variance decomposition. There have been quite a lot of studies conducted for diversity quantification and creation, which have provided useful guidance on how to construct effectively a high quality ensemble leading to the improvement of the generalization performance. However, there is still not yet a formally accepted definition of the term `diversity'~\cite{Kuncheva:Diversity,Zhou:Ensemble,Zhou:Forest}, which indicates that different measures of diversity were designed from different views of diversity and the ensemble diversity was usually created in different heuristic ways~\cite{Zhou:DF}. Moreover, a great number of studies have been conducted towards optimizing the ensemble accuracy and the diversity simultaneously and some works also involve introducing new metrics of diversity quantification towards enhancing heuristically the relationship between the ensemble accuracy and the diversity. However, the above-mentioned relationship still needs to be explored further in depth to make it more clear how the simultaneous optimization of the ensemble accuracy and the diversity can be achieved more effectively. 






\subsection{Research Gaps}
\label{sec:sec:gap}
Despite considerable progress in handling the epistemic uncertainty in supervised learning models, there exist several challenging issues that must be addressed in the future. We found several research gaps that need further investigations, as follows:
\begin{itemize}
    \item \textbf{Methodology}: Although supervised learning approaches have been widely applied to solve computer vision and NLP problems, most of the existing studies fail to quantify uncertainty in practice. They usually use ideal (standard) data sets and inject a uniform random noise to evaluate their performance, which is unrealistic in real-world problems. In practice, the performance of learning from data sets are affected by uncertain distributions; therefore, it is crucial to develop robust techniques for learning uncertainty. Moreover, in NLP, the uncertainty on the contexts of words is naturally present in the text due to the insufficient amount of data but very few studies on epistemic uncertainty have been conducted in this aspect, which indicates the necessity of further studies on uncertainty in text processing. 
    In addition, most of the studies estimated the uncertainty in supervised learning models, while little attention has been paid to other learning strategies such as semi-supervised learning~\cite{pouroanah2021asemisuoervised}, multi-modal learning~\cite{wu2021target}, reinforcement learning~\cite{pourpanah2019animproved}, active learning~\cite{wang2017incorporating}, transfer learning~\cite{shiu2001transferring}, graph learning~\cite{liu2012anew}, etc. 
    {\color{black}From the perspective of algorithm optimization, choosing suitable epistemic uncertainty quantification methods according to specific tasks and algorithm characteristics, and generating an optimized learning strategy based on quantified uncertainty, is worth further exploration. It may be an effective way to improve the performance of deep neural networks with different structural characteristics and other classic learning algorithms. For example, there are several excellent evolutionary computation algorithms, such as particle swarm optimization~\cite{zeng2021competitive,liu2021novel,liu2019novel}, that have received extensive attention from researchers in the post-deep learning era. However, there is little research work on quantifying the uncertainty of such algorithms. We believe that the epistemic uncertainty learning techniques can be used to improve the stability of the optimization process.
    
    \item \textbf{Lack of data set}: For the topic of model uncertainty quantification, there are not yet benchmark databases designed particularly. The data sets used in this study are from areas of CV and NLP. Nonetheless, it is one of the basements for studying epistemic uncertainty quantification. Analyzing epistemic uncertainty based on bias-variance decomposition may be a breakthrough in constructing a benchmark data set that reflects the effectiveness of epistemic uncertainty quantification techniques fairly. We will explore this further in our future work.}
    
    \item \textbf{Lack of standard evaluation protocol}: Existing uncertainty learning techniques are being evaluated based on measurable quantities such as performance on out-of-distribution detection. {\color{black}However, the details of such performance evaluation strategy may vary for different studies, which leads to an unfair comparison among various techniques. Therefore, it is vital to have a standard protocol for evaluating the effectiveness of uncertainty quantification techniques directly. Based on the bias and variance decomposition, which is mentioned in this work, making theoretical exploration of the quality of the uncertainty estimation is also a good future research direction.}

    \item \textbf{Availability of data and code:} This can help researchers to reproduce results, enhance their performance and conduct a fair comparison. However, the majority of studies do not make the relevant codes and data available.
    
\end{itemize}

\color{black}
\subsection{Concluding remarks}
\label{sec:sec:conc}

Enabling supervised learning models to quantify their uncertainty is vital for many real-world applications such as safety-related problems. {\color{black} This survey first explained the importance of addressing the epistemic uncertainty in supervised learning models and discussed it in terms of bias and variance. Then,} we reviewed the epistemic uncertainty learning techniques in supervised learning over the last five years. We provided a hierarchical categorization of these techniques and introduced the representative models of each category along with their applications. Specifically, we discussed two widely used epistemic uncertainty learning techniques, i.e., Bayesian approximation and ensemble learning. In addition, several research gaps have been pointed out as potential future research directions. It is aimed to promote the concept of epistemic uncertainty learning.

\section*{Acknowledgment}
 This work was supported in part by the National Natural
 Science Foundation of China (Grants 61976141, 62176160 and 61732011), in part by National Key R\&D Program of China (Grant 2021YFE0203700), in part by the Natural Science Foundation of Shenzhen (University Stability Support Program no. 20200804193857002), and in part by the Interdisciplinary Innovation Team of Shenzhen University.

\bibliography{mybibfile}

\end{document}